%% file: main.tex
\pdfoutput=1

\documentclass[11pt]{article}

\usepackage[final]{acl}

\usepackage{times}
\usepackage{latexsym}
\usepackage[T1]{fontenc}

\usepackage{svg}
\usepackage[utf8]{inputenc}
\usepackage{xspace}
\usepackage{microtype}
\usepackage{inconsolata}
\usepackage{colortbl}
\usepackage{amsmath}
\usepackage{amssymb}
\usepackage{array}
\usepackage{pifont}
\usepackage{tabularx}
\usepackage{adjustbox}
\usepackage{enumitem}
\usepackage{multirow}
\usepackage{dsfont}
\usepackage{booktabs}
\usepackage{MnSymbol}
\usepackage{scalerel}
\usepackage{mathrsfs}
\usepackage{graphicx}
\usepackage{changes}
\usepackage{xcolor,soul}
\usepackage{makecell}
\usepackage{ulem}
\usepackage{graphicx,calc}
\usepackage{bbm}
\usepackage{lipsum}
\usepackage[caption=false]{subfig}
\usepackage{algorithm}
\usepackage{algpseudocode}
\usepackage{amsthm}

\usepackage[most]{tcolorbox}

\newtcbox{\hlprimarytab}{on line, rounded corners, box align=base, colback=white!10,colframe=white,size=fbox,arc=3pt, before upper=\strut, top=-2pt, bottom=-4pt, left=-2pt, right=-2pt, boxrule=0pt}
\newtcbox{\hlprimarytabg}{on line, rounded corners, box align=base, colback=gray!10,colframe=white,size=fbox,arc=3pt, before upper=\strut, top=-2pt, bottom=-4pt, left=-2pt, right=-2pt, boxrule=0pt}
\newtcbox{\hlsecondarytab}{on line, box align=base, colback=red!10,colframe=white,size=fbox,arc=3pt, before upper=\strut, top=-2pt, bottom=-4pt, left=-2pt, right=-2pt, boxrule=0pt}

\definecolor{darkgreen}{RGB}{0,100,0}
\definecolor{darkred}{RGB}{200,0,0}
\definecolor{lightgreen}{RGB}{228,253,227}
\definecolor{lightred}{RGB}{252,231,234}
\definecolor{lightyellow}{RGB}{250,253,191}
\definecolor{lightblue}{RGB}{230,240,254}
\definecolor{white}{RGB}{255,255,255}

\newcommand{\whethermath}[1]{\ifmmode{#1}\else{$#1$}\fi}

\newcommand{\phz}{\ifmmode\phantom{0}\else$\phantom{0}$\fi}

\usepackage{cleveref}

\newtheorem{proposition}{Proposition}
\newtheorem{lemma}{Lemma}
\newtheorem{claim}{Claim}
\title{Mitigating Coordinate Prediction Bias from Positional Encoding Failures}

\author{Xingjian Tao\textsuperscript{1}, Yiwei Wang\textsuperscript{3}, Yujun Cai\textsuperscript{4}, 
Yihong Luo\textsuperscript{2}, Kai Han\textsuperscript{5}, and Jing Tang\textsuperscript{1,2}\thanks{Corresponding Author: Jing Tang.}\\\\
\textsuperscript{1}The Hong Kong University of Science and Technology (Guangzhou)\\ ~\textsuperscript{2}The Hong Kong University of Science and Technology
~\textsuperscript{3}University of California, Los Angeles
\\~\textsuperscript{4}The University of Queensland
~\textsuperscript{5}Shanghai University of Finance and Economics
\\
\texttt{taoxj2001@outlook.com}\\
}

\begin{document}
\maketitle
\begin{abstract}
While Multimodal Large Language Models (MLLMs) excel at general vision-language tasks, precise coordinate prediction remains a significant challenge, particularly as high-resolution inputs cause visual positional encodings (VPEs) to degrade. 
We demonstrate that these encoding failures do not result in random noise but instead trigger predictable, directional biases, suggesting that models default to internal spatial priors when grounding signals are weak. To counteract this, we introduce Vision-PE Shuffle Guidance (VPSG), a training-free, inference-time correction method. VPSG isolates position-unconditioned tendencies by shuffling VPEs and utilizes this negative evidence to steer digit decoding through a lightweight finite-state machine. Evaluation on the ScreenSpot-Pro benchmark confirms that VPSG effectively rectifies coordinate drift, yielding consistent improvements in localization accuracy across various model scales without any retraining. Our code is available at 
\href{https://github.com/taoxj2001/VPSG}{https://github.com/taoxj2001/VPSG}

\end{abstract}

\input{sec/1introduction}
\input{sec/2related}
\input{sec/3method}

\input{sec/4experiment}

\input{sec/5conclusion}

\section*{Limitations}
This study has several limitations. We evaluated only a small number of models and did not include larger parameter sizes. The current evaluation method does not account for incorrectly formatted or invalid coordinate outputs. Furthermore the approach is restricted to English. Expansion to other languages remains in the early stages of development

\section*{Acknowledgments}
This work is partially supported by National Key R\&D Program of China under Grant No. 2023YFF0725100, by the National Natural Science Foundation of China (NSFC) under Grant No. 62402410, by Guangdong Provincial Project (No. 2023QN10X025), by Guangdong Basic and Applied Basic Research Foundation under Grant No. 2023A1515110131, by Guangzhou Municipal Education Bureau (No. 2024312263), by Nansha District Project (No. 2023ZD022), by HKUST(GZ) Kunpeng\&Ascend Center of Cultivation, and by the Fundamental Research Funds for the Central Universities.

\bibliography{main}

\appendix
\clearpage
\input{sec/appendix}

\end{document}

%% file: sec/1introduction.tex
\section{Introduction}

\begin{figure*}[t!]
    \centering
    \includegraphics[width=\linewidth]{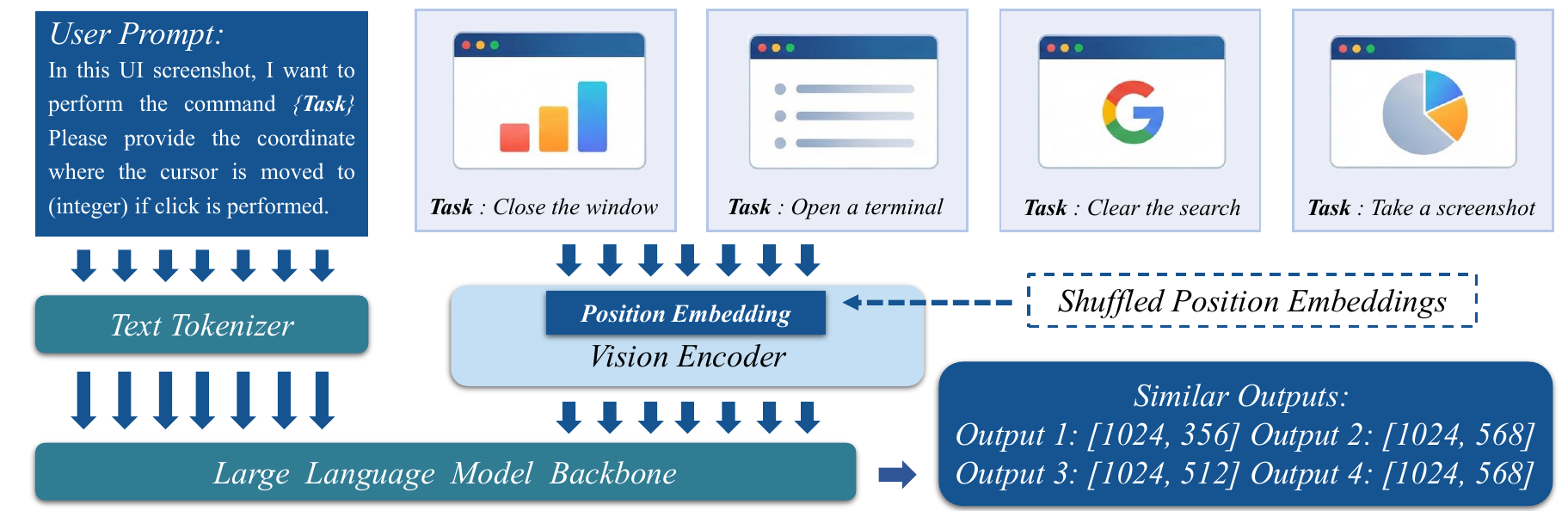}
    \caption{Effect of shuffling visual positional encodings: removing spatial conditioning causes the model to collapse to similar coordinate predictions across independent runs, indicating a position-unconditioned, directional bias rather than random variation. We also observe a similar clustered error pattern on high-resolution images (without shuffling), consistent with position-encoding failures.}
    \label{fig:intro}
\end{figure*}

Recent advances in large language models (LLMs; \citealt{touvron2023llama1,vicuna2023,falcon40b,MosaicML2023Introducing,touvron2023llama,openaichatgptblog,bardclaudeblog}) have improved language understanding and generation, but their text-only I/O limits perceptual and interactive use. 
Multi-modal LLMs (MLLMs) combine vision and text, e.g., Flamingo~\citep{alayrac2022flamingo}, Gemini~\citep{team2023gemini}, and Qwen-VL~\citep{bai2023qwen,wang2024qwen2,bai2025qwen25vltechnicalreport} to enable tasks such as visual QA, captioning, and document understanding.  
Coordinate prediction supports applications like object manipulation and GUI automation, where an MLLM outputs a 2-D point or bounding box (e.g., “[1000,500]”). High-resolution inputs make this harder: token and compute costs rise, and larger spatial extents increase pixel–patch misalignment, leading to coordinate drift~\citep{li2025screenspot, gao2024insights, hsieh2024found, yen2024longcontextlanguagemodelingparallel}.

Coordinate drift originates from the failure of conventional positional encodings (PEs) to scale reliably at high resolutions~\citep{zhang2024beyond}. Although essential for anchoring visual tokens to image geometry , standard schemes—typically 2-D encodings in the vision encoder~\citep{dosovitskiy2020image} and sequence-level RoPE~\citep{su2024roformer} in the LLM —suffer as high-resolution inputs induce long-context regimes that diffuse attention and weaken spatial cues. Existing remedies are insufficient for precise localization: cropping-based methods~\citep{tao2025understanding, wu2025dimo} often compromise global semantics , while PE enhancements~\citep{ge2024v2pe, chen2025advancing, heo2024rotary} address coarse tasks like VQA but fail to mitigate the subtle biases that drive numerical coordinate errors.

Coordinate prediction requires precise spatial awareness; however, unlike humans who resort to random guessing when positional signals are shuffled, MLLMs exhibit predictable \textit{directional biases}. These patterns closely mirror error distributions in natural high-resolution datasets, suggesting that the weakening of positional information at scale amplifies the model's inherent priors and degrades coordinate accuracy as shown in~\Cref{fig:intro}. In this work, we seek to mitigate these biases and reinforce positional encodings to improve MLLM robustness on position-sensitive tasks.

We propose Vision-PE Shuffle Guidance (VPSG), a training-free, test-time method that rectifies coordinate biases online using only the base model. VPSG contrasts a position-conditioned main route with auxiliary routes that approximate a position-unconditioned reference by shuffling visual positional encodings. This fusion amplifies spatially grounded information while suppressing spurious content that persists even when positional cues are removed. Practically, a lightweight FSM ensures that only digit tokens are adjusted—down-weighting position-free tendencies while leaving formatting characters like commas and brackets untouched. This mechanism functions as a scaled ``negative score'' during decoding, effectively reinforcing positional influence and stabilizing $[x,y]$ outputs without requiring any training or architectural modifications.

Two key design choices make VPSG precise and stable. Rather than relying on a single PE-shuffled auxiliary route, VPSG aggregates multiple shuffled routes in log space (geometric mean), yielding a robust estimate of the position-unconditioned bias and stabilizing the negative-evidence signal across inputs. In addition, VPSG applies a position-aware coefficient schedule: the guidance weight starts high for the first digit of $x$, decays geometrically for subsequent digits, resets at the first digit of $y$, and then decays again. This concentrates correction on the most influential digits while avoiding over-regularization of later positions and preserving natural numeric formatting.
Our contributions can be summarized as follows:

\begin{itemize}[topsep=4pt, itemsep=0pt]
    \item \textbf{Positional fragility analysis.} We show that perturbing visual positional encodings (VPEs) induces directional, repeatable biases in MLLM coordinate prediction, with similar effects at high resolution—revealing a resolution-dependent failure mode.
    \item \textbf{Training-free guidance.} We propose VPSG, a model-agnostic test-time method that shuffles VPEs to form counterfactual routes and guides final-layer digit logits via a lightweight finite-state machine.
\end{itemize}

%% file: sec/2related.tex
\section{Related Work}
\paragraph{Coordinate prediction with MLLMs}
With the advancement of MLLMs \citep{bai2023qwen, wang2024qwen2, bai2025qwen25vltechnicalreport, alayrac2022flamingo, team2023gemini, ma2023llm, yang2023dawn, liu2023visual, li2023llava, liu2023llavapluslearningusetools, xu2025qwen25omnitechnicalreport}, coordinate prediction has shifted to language-driven grounding, where coordinates are generated as discrete token sequences. This paradigm is vital for GUI interaction, supported by specialized datasets designed for screenshot-based grounding \citep{cheng2024seeclick, wu2024atlas, li2025screenspot}. To enhance UI navigation and agent autonomy, data-centric methods such as ShowUI \citep{lin2024showui} and UGround \citep{gou2025uground} utilize large-scale synthetic datasets. Furthermore, specialized frameworks like OmniParser \citep{lu2024omniparser} leverage auxiliary visual models to improve element localization , while WebGUM \citep{furuta2023multimodal} integrates hierarchical planning with perceptual grounding to enable structured web-based decision-making.

\paragraph{Visual position encoding}
Visual position encoding is essential for scaling multimodal transformers \citep{wang2025eliminatingpositionbiaslanguage}. Variable schemes like V2PE \citep{ge2024v2pe} and PyPE \citep{chen2025advancing} improve robustness in long context perception. Lightweight 2D markers enhance layout understanding in structured document tasks \citep{liao2025doclayllm}. RoPE provides extrapolation benefits and semantic aware encodings adapt to perceptual similarity \citep{heo2024rotary, chen20252d}. Recent studies generalize RoPE through Fourier analysis \citep{hua2024fourier} or trainable matrices \citep{yu2025comrope} and explore pooling interactions \citep{lee2025maximizing}. Large vision embedding norms can suppress positional information and cause spatial reasoning failures \citeauthor{qi2025beyond} \citep{qi2025beyond}. These results show that robust encodings are central to model advancement. However research on tasks requiring precise location information like coordinate prediction remains scarce.

%% file: sec/3method.tex
\section{Methodology}

In this section, we introduce Vision-PE Shuffle Guidance (VPSG), a training-free, test-time guidance that stabilizes coordinate outputs in multimodal LLMs. VPSG runs the base model once on the normal input and in parallel creates shuffle-guided auxiliary views by perturbing only the visual positional encodings; discrepancies across these views reveal spurious numeric tendencies. Using this as negative evidence, VPSG gently steers the main decoding on digit tokens (leaving non-digits untouched) and improves $[x,y]$ reliability without fine-tuning or architectural changes.

\subsection{Causal view of coordinate prediction}

\begin{figure}
  \begin{center}
    \includegraphics[width=1\linewidth]{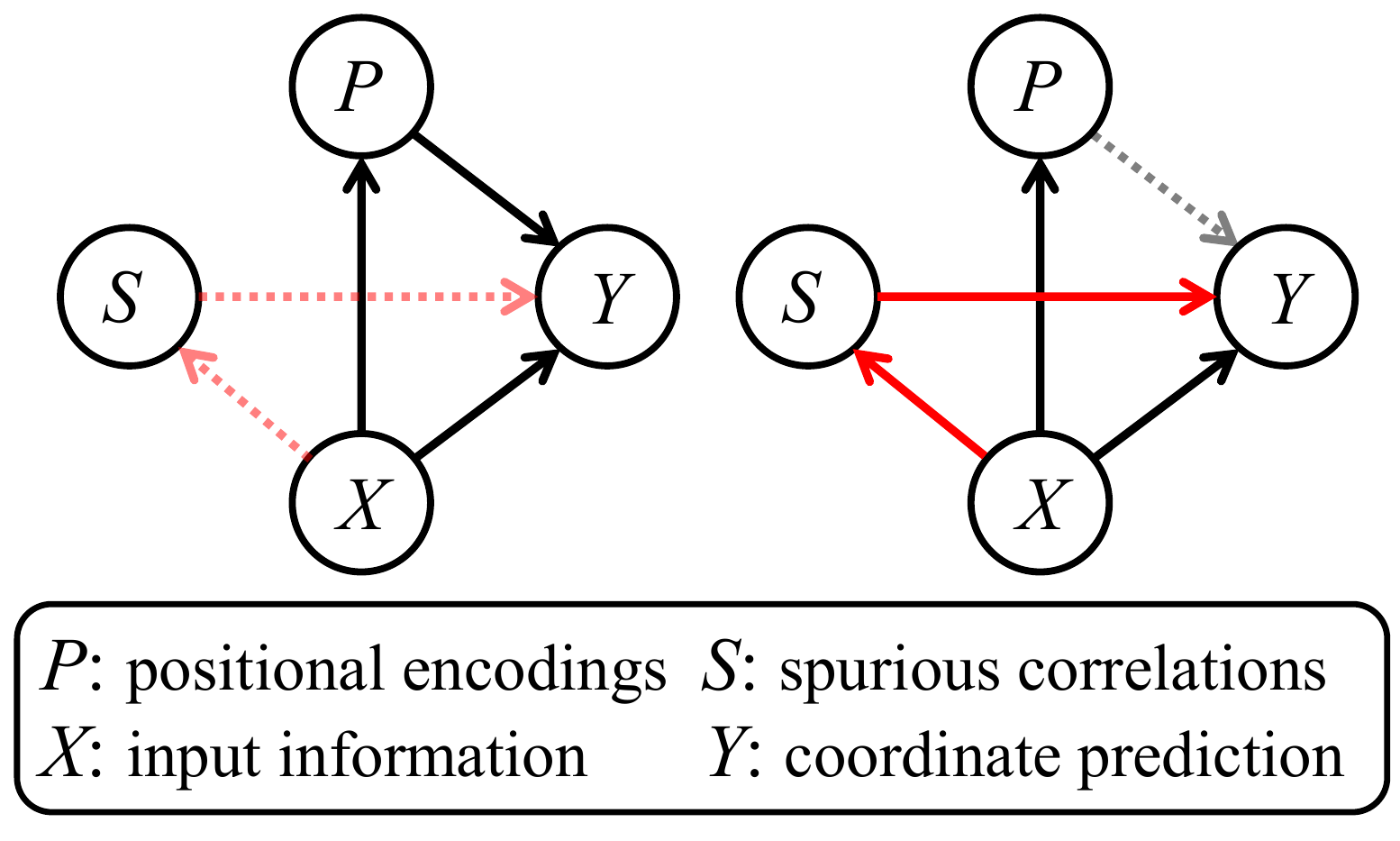}
  \end{center}
  \caption{Causal view of coordinate prediction. Image content and prompt provide the intended causal effect on output digits (left). When VPEs are missing (e.g., at high resolution), the model relies on spurious correlations, leading to directional digit biases (right).}
\label{fig:causality}
\end{figure}

At the core of our study lies the observation that coordinate prediction in multimodal LLMs is governed by a causal mechanism: the output depends simultaneously on position-conditioned signals (e.g., visual positional encodings) and position-unconditioned signals (e.g., default digit tendencies). Prior analyses largely overlook the influence of these position-independent inputs, treating prediction errors as random noise. In contrast, we hypothesize that the non-positional pathway can introduce directional bias when positional information is weak or missing. Motivated by this insight, we adopt a causal graph~\citep{pearl2016causal, pearl2018causal, tang2020unbiased, wang2022should} as the analytical framework to explicitly model how position-related and position-free factors jointly shape the output and to identify spurious routes that degrade coordinate accuracy. 
Ideally, the output digits $[x,y]$ are determined jointly by the image content and the textual prompt, with visual positional encodings (VPEs) supplying spatial grounding. 
However, when VPEs are missing or unreliable—such as in high-resolution inputs, the causal pathway is disrupted. Lacking accurate positional cues, the model tends to rely on spurious correlations, for example overpredicting certain digits or repeating biased numeric patterns that are not grounded in the image. 

As shown in~\Cref{fig:causality}, 
we model coordinate prediction with a simple structural view over four nodes: input information $X$ (image content and prompt), positional encodings $P$ (visual spatial cues), spurious correlations $S$ (default numeric tendencies that emerge independently of the input, such as frequently repeated digits or preferred coordinate patterns), and the predicted coordinates $Y$. 
Here, $S$ captures position-unconditioned regularities in the model’s training data or internal priors that can influence outputs even when visual evidence is weak or missing.
A minimal Structural Causal Model is
$Y = g(X,P,S),$ where $P$ is provided by the vision encoder (and varies with resolution), and $S$ summarizes non-causal numeric regularities the model can fall back on.

When $P$ is available and reliable, it supplies spatial grounding. The dominant pathways are $X\!\to\!Y$ and $P\!\to\!Y$. The influence of $S$ is negligible (dotted edges): although spurious patterns exist in the model’s priors, they are largely blocked in practice because the model can rely on informative $X$ and $P$. 
When $P$ is absent or unreliable, the informative pathway weakens ($P\!\dashrightarrow\!Y$), and the spurious path without positional condition $S\!\to\!Y$ becomes comparatively strong. The model then defaults to biased numeric templates (e.g., over-predicting certain digits or repeating patterns) that are not supported by the input. In potential-outcome terms, the discrepancy
$S(x) \;=\; Y(x,p_{\text{bad}})\;-\;Y(x,p_{\text{good}})$
captures the shift in predictions attributable to the loss of positional grounding, with $p_{\text{good}}$ denoting a reliable PE setting and $p_{\text{bad}}$ an out-of-range or missing one.

This causal view clarifies the failure mode: positional degradation amplifies the non-causal route from $S$ to $Y$, yielding directional digit errors even when $X$ is unchanged. It also motivates our methodology: design a test-time procedure that (i) exposes the spurious route when $P$ is weak and (ii) suppresses its influence on the numeric tokens while preserving the informative flow from $X$ (and any usable $P$).


\subsection{Bias analysis}

Given the causal graph in~\Cref{fig:causality}, weakening the positional encodings $P$ increases the relative influence of spurious correlations $S$ on the output $Y$, distorting the digit distribution and pulling predictions away from the evidence in $X$. We therefore propose a two-part intervention: (i) expose the spurious route when $P$ is weak by constructing counterfactual views that differ only in positional cues, and (ii) suppress its impact on numeric tokens while preserving the informative flow from $X$ (and any usable $P$). Before presenting the intervention, we first establish empirically that the resulting errors are directional rather than random, confirming that the $S\!\to\!Y$ pathway is measurable.

Let $x$ be an input of size $(W_x,H_x)$ with diagonal $d_x=\sqrt{W_x^2+H_x^2}$ to normalize scale across images.  
Under shuffled PEs, we run the model on $S$ cases and compute pairwise distances
\begin{equation}
d^{(i,j)}(x)=\big\|\hat{\mathbf y}^{(i)}(x)-\hat{\mathbf y}^{(j)}(x)\big\|_2,
\qquad 1\le i<j\le S,
\end{equation}
then normalize as
$\tilde d^{(i,j)}(x)=\frac{d^{(i,j)}(x)}{d_x}$.
Pooling $\tilde d^{(i,j)}(x)$ over all inputs yields the shuffled-PE distance distribution $\mathcal{P}_{\text{shuffle}}$, while normal PEs give $\mathcal{P}_{\text{normal}}$. 
We compare their empirical means, against the scale-aware baseline $\mu_0\approx0.5214$ (Appendix~\ref{app:proof1}) to assess whether shuffled predictions collapse toward a small coordinate subset.
Evidence of systematic, non-random bias is a clear left shift of $\mathcal{P}_{\text{shuffle}}$ toward zero relative to both $\mu_0$ (i.e., $\mathbb{E}[\tilde d]\ll \mu_0$) and $\mathcal{P}_{\text{normal}}$, indicating that predictions under shuffled PEs collapse to a few favored coordinates rather than dispersing as random fluctuations would.

\begin{figure*}[htbp]
\centering
\begin{minipage}[t]{0.31\textwidth}
  \centering
  \includegraphics[width=\linewidth]{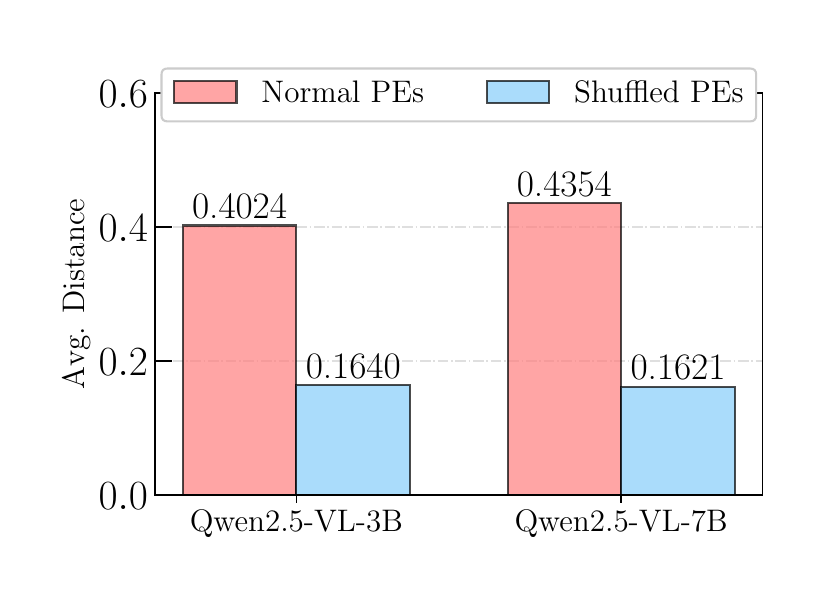}
  \caption{
  Diagonal-normalized average pairwise distance $\tilde d$ between coordinate predictions under \emph{Normal PEs} and \emph{Shuffled PEs}. 
  }
  \label{fig:distance_bar}
\end{minipage}
\hfill
\begin{minipage}[t]{0.65\textwidth}
  \centering
  \includegraphics[width=\linewidth]{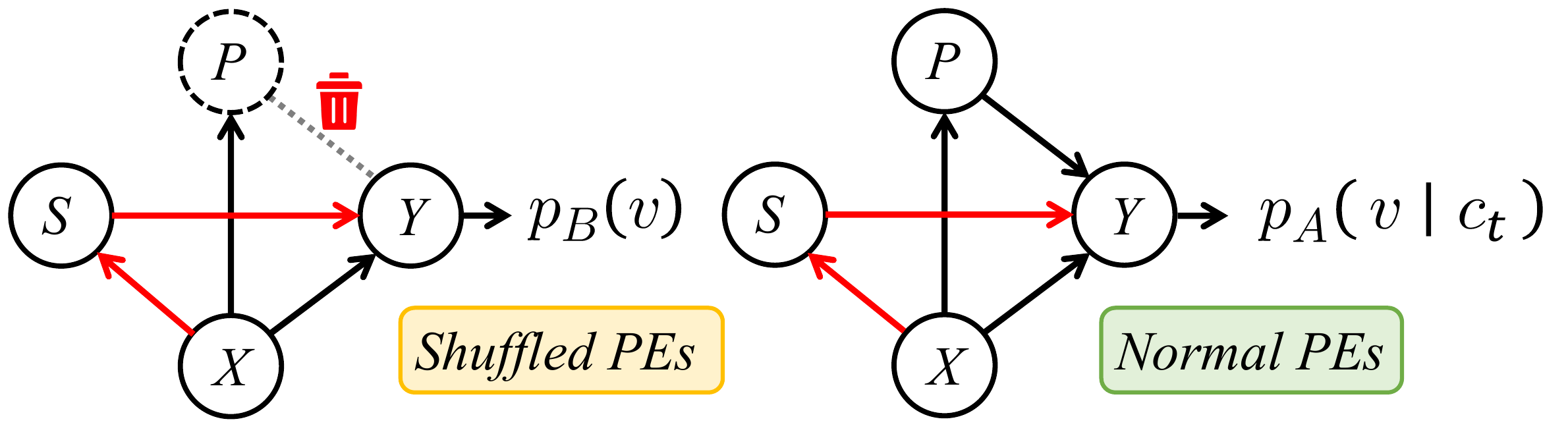}
  \caption{
  We obtain a position\textit{-unconditioned} reference by shuffling visual positional encodings (left) and fuse it with the position\textit{-conditioned} prediction (right). 
  This conditional–unconditional contrast serves as negative evidence for digits, strengthening positional cues and suppressing spurious numeric patterns during decoding.
  }
  \label{fig:method}
\end{minipage}
\end{figure*}

\Cref{fig:distance_bar} summarizes the distance statistics. Across both Qwen2.5-VL-3B and Qwen2.5-VL-7B, the diagonal-normalized average pairwise distance under shuffled positional encodings is consistently small (\(\tilde d \approx 0.16\)),
whereas the normal-PE condition exhibits substantially larger dispersion (\(\tilde d \approx 0.40\)–\(0.44\)).
This substantial gap, far exceeding the baseline dispersion of random uniform points, confirms that when positional encodings are disrupted the model outputs collapse to a small set of preferred coordinates rather than spreading randomly.   The consistent pattern across model scales demonstrates that the observed systematic directional bias is an inherent property of the architecture rather than a size-related artifact.

Based on the empirical and causal analysis of systematic bias caused by missing positional encodings presented above, we propose a bias-reduction strategy that remains unchanged during training but is inserted during inference: Vision-PE Shuffle Guidance (VPSG). Inspired by classifier-free guidance (CFG)~\citep{ho2022classifier}, we mitigate the impact of directional bias from the perspective of probability distribution. 

\paragraph{Overall algorithm.}

VPSG runs one \emph{main route} with normal visual positional encodings (PEs) and several \emph{auxiliary routes} with randomly shuffled PEs. We let $c_t$ denote the position-conditioned context at step $t$, which primarily captures the positional encoding information from the visual encoder that provides spatial grounding for decoding.
At each decoding step, we contrast the \textbf{position-conditioned} prediction $p_A(v\mid c_t)$ from the main route with an aggregated, \textbf{position-unconditioned} reference $p_B(v)$ formed by combining multiple shuffled routes. 
This contrast acts as \emph{negative evidence} on digit tokens, while non-digit tokens (commas, spaces, brackets) remain untouched as shown in~\Cref{fig:method}. 
A finite-state machine (FSM) tracks whether the model is decoding the $x$ or $y$ coordinate and indicates when digit-specific guidance should be applied.  
This mechanism preserves the required $[x,y]$ format throughout decoding and prevents structural errors that could arise from spurious tokens or misaligned guidance.

\begin{proposition}[VPSG token guidance]
Let $\mathcal{D}\subset\mathcal{V}$ denote the digit subset of the vocabulary and $\alpha_t$ the step-wise guidance coefficient determined by the FSM.  
The VPSG-adjusted distribution satisfies
\begin{equation}
\begin{aligned}
&p_{\mathrm{VPSG}}(v\mid c_t) \propto \\
&\begin{cases}
\exp\bigl(\log p_A(v\mid c_t)-\alpha_t\tilde{\ell}_B(v)\bigr), & v\in \mathcal{D},\\[3pt]
p_A(v\mid c_t), & v\notin \mathcal{D},
\end{cases}
\end{aligned}
\end{equation}
where $\tilde{\ell}_B(v)=\log p_B(v)$ is the log-probability of the position-unconditioned reference.
\end{proposition}

This compact formula shows that VPSG subtracts a scaled log-probability from the digit logits of the main route while leaving non-digit tokens unchanged.  
The aggregation of $p_B(v)$ and the scheduling of $\alpha_t$ are described below; the proof that this form is equivalent to the classifier-free guidance view is provided in Appendix~\ref{app:vpsg-proof}.

\begin{figure*}[t!]
    \centering
    \includegraphics[width=\linewidth]{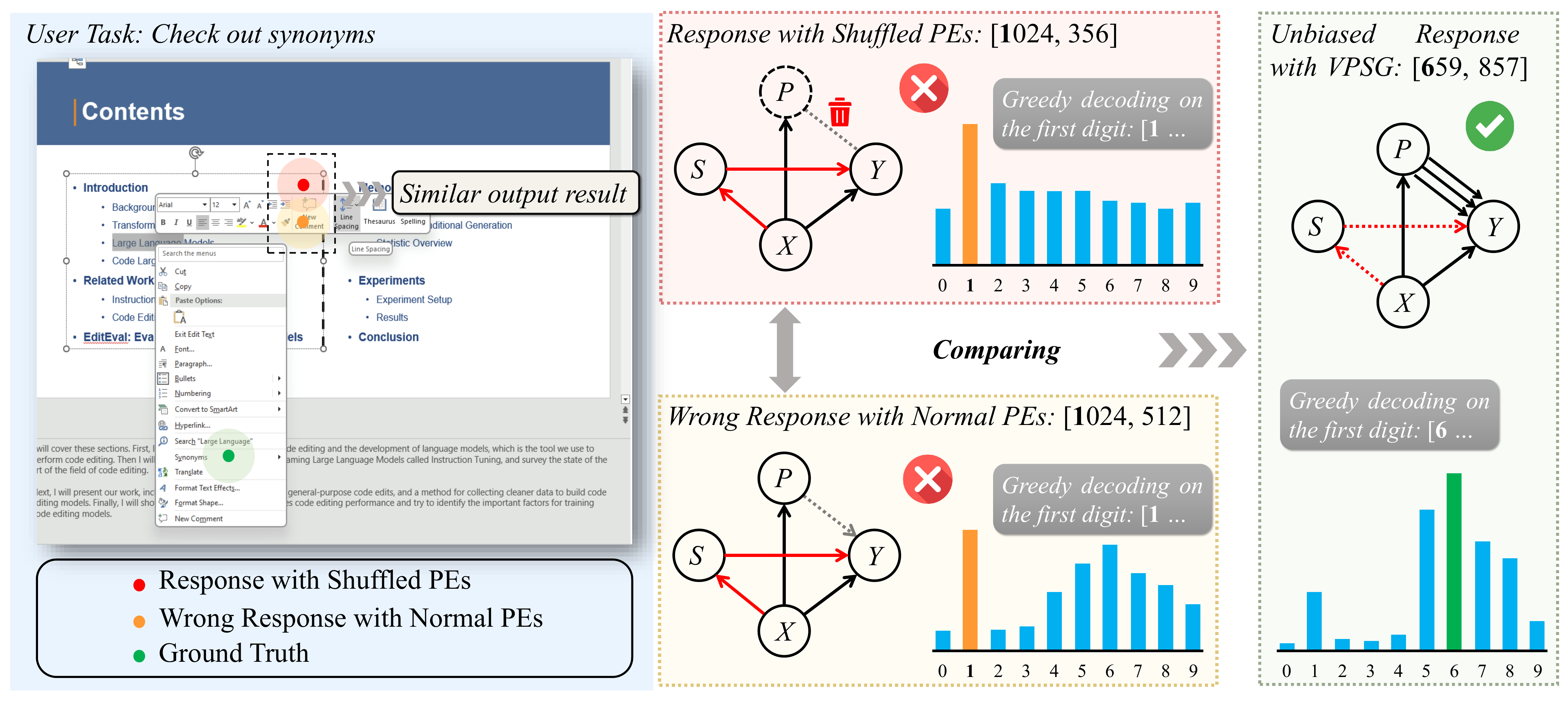}
    \caption{
Qualitative example of VPSG on a Screenspot-Pro case.  
The base model with normal positional encodings produces a biased coordinate prediction ([1024, 512]), while the same model with shuffled positional encodings collapses to a similar but consistently biased point ([1024, 356]), revealing a directional position-unconditioned tendency.   Applying VPSG corrects this bias and outputs the accurate ground truth ([659, 857]) by integrating negative evidence from multiple shuffled runs and reweighting digit logits, demonstrating how VPSG suppresses spurious patterns and restores faithful spatial grounding. 
    }
    \label{fig:case}
\end{figure*}

\paragraph{Seeds aggregation.}
The auxiliary reference $p_B(v)$ is estimated by running the model on $S$ independent PE-shuffled seeds and aggregating their log-probabilities,
\begin{multline}
\tilde{\ell}_B(v) = \mathbb{E}_{\sigma\sim\nu}\!\left[\,\log p_B^{(\sigma)}(v)\,\right] \\
= \int_{\sigma\in\mathcal{S}} \log p_B^{(\sigma)}(v)\, d\nu(\sigma) \approx \frac{1}{S}\sum_{s=1}^{S} \log p_B^{(s)}(v),
\end{multline}
where $\sigma$ indexes shuffle transformations and $\nu$ is the (uniform) seed measure.
This log-space (geometric mean) aggregation provides a robust Monte-Carlo estimate of the position-unconditioned bias prior.

\paragraph{Coefficient decay.}
To focus correction on the most influential digits, VPSG uses a geometric decay on $\alpha_t$ along the decoding sequence.
Let $k_x$ (resp.\ $k_y$) be the index of the current digit within $x$ (resp.\ $y$).  
The guidance coefficient is scheduled as
\begin{equation}
\alpha_t=
\begin{cases}
\alpha\,\mathrm{decay}^{\,k_x-1}, & \text{the }k_x\text{-th digit of }x,\\[2pt]
\alpha, & \text{first digit of }y,\\[2pt]
\alpha\,\mathrm{decay}^{\,k_y-1}, & \text{the }k_y\text{-th digit of }y,
\end{cases}
\end{equation}
where $0<\mathrm{decay}<1$.
This schedule emphasizes the most significant digits, resets at the first $y$ digit, and then tapers off, preventing over-regularization on later positions.

\paragraph{Summary.}
Overall, VPSG addresses this by effectively “asking twice”: once with the normal input and again with a version whose visual positions are shuffled. Disagreements from the shuffled view act as negative evidence, gently steering the model’s digit choices back toward what the image supports, while leaving non-digit text untouched. The result is a simple, plug-in procedure at inference time that stabilizes $[x,y]$ predictions without changing training or model architecture.
It is model-agnostic, requires no retraining, and exactly recovers the baseline behavior when $\alpha_t\to 0$. The complete VPSG algorithm is shown in~\Cref{app:algo}. As shown in~\Cref{fig:case}, we provide a qualitative example of VPSG on a Screenspot-Pro case.

%% file: sec/4experiment.tex
\section{Experiments}

In this section, we evaluate the performance of VPSG when applied to VLLMs. As a plug-and-play, training-free approach, VPSG enhances existing models on coordinate prediction tasks, providing consistent improvements without requiring additional fine-tuning or architectural modifications.

\subsection{Experimental settings}
\paragraph{Datasets.} We adopt the widely used ScreenSpot-Pro dataset to evaluate the performance of our method. ScreenSpot-Pro is a recently released benchmark for GUI grounding, consisting of real high-resolution desktop screenshots spanning 23 applications (e.g., VSCode, Photoshop, AutoCAD), five industry categories, and three operating systems, with precise annotations provided by professional users as shown in~\Cref{app:dataset}. The dataset is particularly challenging because target UI elements are often extremely small, occupying on average only 0.07\% of the screen area. We evaluate VPSG under this realistic, high-resolution, and difficult setting to validate its effectiveness in improving localization accuracy.


\paragraph{Models.} We adopt Qwen2.5-VL~\citep{bai2025qwen25vltechnicalreport} as our test model, including configurations with 3B and 7B parameters. Unlike previous multimodal models, Qwen2.5-VL can directly output absolute coordinates for grounding tasks without requiring additional post-hoc alignment. The Qwen2.5-VL series has demonstrated strong performance on standard coordinate prediction benchmarks, even surpassing some specialized GUI models. However, its performance still degrades considerably in high-resolution scenarios, highlighting the inherent difficulty of precise localization under long-context inputs.
\paragraph{Method configurations.} For all evaluated models, we adopt greedy decoding to eliminate randomness and ensure reproducibility. 
For our proposed method VPSG, we identify the optimal hyperparameter configuration through grid search, 
with $\alpha = 0.55$ and $\text{decay} = 0.4$.
\paragraph{Compared models}
We evaluate a broad spectrum of multimodal models with
an emphasis on general-purpose and training-free baselines, which are particularly important for assessing the effectiveness of our method without additional fine-tuning.
This group includes Qwen-VL-7B~\citep{bai2023qwen}, GPT-4o~\citep{achiam2023gpt}, Qwen2-VL-7B~\citep{wang2024qwen2}, and MiniCPM-V, representing strong
generalist vision–language models that can directly perform coordinate prediction. We further include the recent Qwen2.5-VL family (3B and 7B), which serves as our primary base model for applying VPSG and can output absolute coordinates without post-hoc alignment.
For completeness, we also report results of specialized GUI action models such as SeeClick~\citep{cheng2024seeclick}, OS-Atlas-4B/7B~\citep{wu2024atlas}, ShowUI-2B~\citep{lin2024showui}, CogAgent~\citep{hong2024cogagent}, Aria-GUI~\citep{yang2024aria}, and UGround-7B~\citep{gou2025uground}, which are trained or
instruction-tuned specifically for interface grounding tasks.
This diverse set of baselines enables a comprehensive evaluation of VPSG across both generic and domain-specific settings.
The experimental results of the above model are cited from~\cite{li2025screenspot}.

\subsection{Overall performance}
\input{sec/table/VPSG}

As shown in~\Cref{tab:main}, VPSG consistently improves both base models on the Screenspot-Pro benchmark when measured by percentage correct. On Qwen2.5-VL-3B, the overall percentage correct increases from 11.6 to 13.3, a gain of 1.7 percentage points. Clear improvements appear in multiple text-oriented categories, including Development (Text) from 18.8 to 24.7 (+5.9 points), Creative (Text) from 16.7 to 20.2 (+3.5 points), CAD (Text) from 8.1 to 10.2 (+2.1 points), Office (Text) from 24.3 to 26.6 (+2.3 points)), and Scientific (Text) from 20.8 to 21.5 (+0.7 points). Several icon-oriented settings also benefit; for example, Development (Icon) rises from 1.4 to 2.1 (+0.7 points), Office (Icon) rises from 1.9 to 5.7 (+3.8 points) indicating that mitigating position-induced bias can stabilize landmark selection even when the target is an icon rather than text.

For Qwen2.5-VL-7B, the overall percentage correct increases from 18.5 to 19.1 (+0.6 points). Notable gains include Development (Text) from 37.7 to 40.9 (+3.2 points) and Office (Text) from 41.8 to 43.5 (+1.7 points), along with improvements in icon-oriented cases such as Office (Icon) from 11.3 to 13.2 (+1.9 points). Taken together, these results show that test-time negative-evidence guidance yields reliable lifts across model scales and interaction modes, enhancing both text-oriented and icon-oriented behaviors by suppressing spurious effects that emerge when positional signals are unreliable.

Our results underscore that a causal analysis of error pathways is essential for effectively mitigating coordinate prediction bias. By explicitly contrasting a position-conditioned distribution—obtained from normal positional encodings—with a position-unconditioned reference—derived from shuffled positional encodings—VPSG highlights and strengthens the influence of positional information in the final token distribution, while suppressing the position-agnostic tendencies that drive systematic, directional errors. 

This comparison clarifies how positional cues causally affect output coordinates and ensures that guidance is grounded in a measurable contrast rather than heuristic adjustment. Because the intervention operates solely on the final-layer logits at test time, it remains fully compatible with pretrained MLLMs and introduces no additional training cost, architectural changes, or data requirements. Consequently, VPSG serves as a model-agnostic, plug-in method applicable to a broad range of coordinate-prediction tasks, enabling consistent and reproducible improvements across datasets and resolutions without modifying the existing training pipeline.

\subsection{Ablation study}
\input{sec/table/ablationSeed}

We perform ablations on Screenspot-Pro (percentage correct) to quantify the contribution of each VPSG component while holding all other settings fixed (same base model and decoding strategy).

\subsubsection{seeds aggregation} 
The first component is seeds aggregation: instead of relying on a single PE-shuffled auxiliary route, the full method aggregates multiple routes in log space (geometric mean). Removing this component (w/o seeds aggregation) leads to a drop in performance. The rationale is that, although the errors induced by missing positional encodings are directional, any single random shuffle yields only a noisy sample from the underlying bias distribution and may not be representative. Aggregating across multiple seeds provides a more faithful estimate of the expectation of this position-unconditioned bias prior. 
This multi-path aggregation better recovers the output distribution absent positional conditioning.

\subsubsection{Coefficient decay}
The second key component of VPSG is coefficient decay. 
Table~\ref{tab:ablation} highlights the importance of this design: 
removing coefficient decay (\emph{w/o coefficient decay}) reduces the average percentage correct from 
13.3 $\rightarrow$ 11.9 (\(\downarrow\) 1.4) on Qwen2.5-VL-3B and from 
19.1 $\rightarrow$ 18.2 (\(\downarrow\) 0.9) on Qwen2.5-VL-7B. 
These drops are substantially larger than those caused by removing seeds aggregation, 
underscoring that position-aware scheduling is a primary driver of VPSG’s gains.
Beyond its positional-error weighting, coefficient decay also compensates for 
confidence attenuation along the digit sequence.
Empirically, we observe that when the model predicts a multi-digit coordinate such as 
\([1234, 567]\), the confidence (logit margin between the top token and the runner-up) 
for the first digit is typically higher than for later digits:
This progressive narrowing of logit margins indicates that later tokens are  intrinsically more ambiguous, making them more sensitive to over-regularization.
Applying a constant guidance coefficient would over-penalize these low-confidence positions, 
potentially distorting fine-scale digits or even the \([x, y]\) template.

By geometrically decaying $\alpha_t$ and resetting at the start of the $y$ coordinate,
VPSG aligns the guidance strength with both positional importance and intrinsic confidence: 
it strongly constrains the high-order digits that dominate absolute error, while reducing the weight 
where the model’s own uncertainty is higher and logit gaps are small.
This targeted scheduling suppresses position-unconditioned biases without compromising the natural 
fine-grained structure of the output.

Taken together, these analyses confirm that coefficient decay is essential for balancing guidance strength with positional and confidence-based considerations, enabling VPSG to suppress position-unconditioned biases effectively.

\subsection{Inference Efficiency}
\label{subsec:efficiency}

Beyond accuracy, we evaluate the practical inference overhead introduced by VPSG on ScreenSpot-Pro benchmark.
Since VPSG runs $S$ auxiliary routes in addition to the original route, the naive theoretical cost would be approximately $(S+1)\times$.
However, as shown in \Cref{tab:efficiency_screenspotpro}, the measured overhead in our setup is substantially smaller:
with $S=3$, VPSG increases the per-case latency from 3.64s to 5.27s on Qwen2.5-VL-3B and from 4.20s to 5.50s on Qwen2.5-VL-7B,
corresponding to only about $1.3\times$--$1.45\times$ slowdown.
These results indicate that VPSG is practically feasible as a plug-in inference-time module for high-resolution GUI grounding pipelines,
providing accuracy gains with a modest latency increase.

\begin{table}[t]
  \centering
  \scriptsize
  \begin{tabular}{lcc}
  
    \toprule
    Setting & Total Time (s) & Time per Case (s) \\
    \midrule
    Baseline (No VPSG, 3B model) & 5755 & 3.64 \\
    Baseline (No VPSG, 7B model) & 6640 & 4.20 \\
    VPSG ($S{=}3$, 3B model)     & 8332 & 5.27 \\
    VPSG ($S{=}3$, 7B model)     & 8696 & 5.50 \\
    \bottomrule
  \end{tabular}
  \caption{Inference efficiency comparison on ScreenSpot-Pro.}
  \label{tab:efficiency_screenspotpro}
\end{table}

We attribute the favorable accuracy--latency trade-off to two key design choices.
First, VPSG is a lightweight, inference-time control that operates only on the final-layer logits and does not require any retraining or architectural modification.
Second, VPSG is strictly activated by a finite-state machine (FSM) and only intervenes when decoding digit tokens (i.e., coordinate digits).
For the majority of output tokens (e.g., brackets, commas, spaces, and other non-digit text), decoding proceeds exactly as in the baseline without guidance.
Together, these design choices keep the additional computation limited, yielding a moderate latency overhead while retaining the benefits of multi-route guidance.

%% file: sec/table/VPSG.tex
\begin{table*}[t!]

 \vspace{5pt}
    \centering
    \setlength{\aboverulesep}{0pt}
    \setlength{\belowrulesep}{0pt}
    \renewcommand{\arraystretch}{1.2}

    \resizebox{1,0\textwidth}{!}{%
        \begin{tabular}{lccccccccccccc}
        \toprule
        \multirow{2}{*}{\textbf{Model}} & \multicolumn{2}{c}{\textbf{Development}} & \multicolumn{2}{c}{\textbf{Creative}} & \multicolumn{2}{c}{\textbf{CAD}} & \multicolumn{2}{c}{\textbf{Scientific}} & \multicolumn{2}{c}{\textbf{Office}} & \multicolumn{2}{c}{\textbf{OS}} & \multirow{2}{*}{\textbf{Avg}} \\
        \cmidrule(lr){2-3} \cmidrule(lr){4-5} \cmidrule(lr){6-7} \cmidrule(lr){8-9} \cmidrule(lr){10-11} \cmidrule(lr){12-13} 
         & \textbf{Text} & \textbf{Icon} & \textbf{Text} & \textbf{Icon} & \textbf{Text} & \textbf{Icon} & \textbf{Text} & \textbf{Icon} & \textbf{Text} & \textbf{Icon} & \textbf{Text} & \textbf{Icon} &  \\

        \midrule
        \multicolumn{12}{l}{\textbf{Trained GUI Action Models}   \,    \raisebox{-0.3ex}{\includegraphics[height=1em]{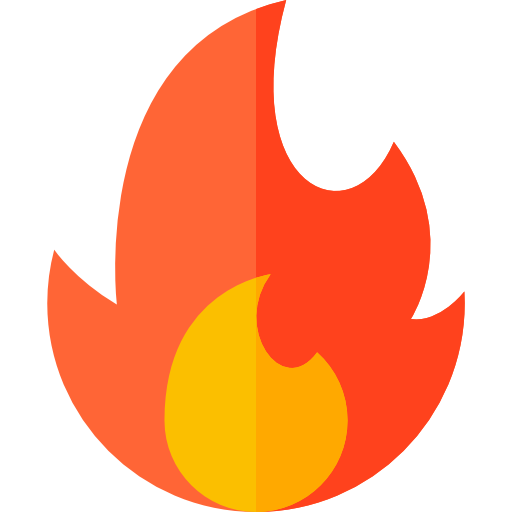}}} \\
        \midrule
        SeeClick         & 0.6  & 0.0  & 1.0  & 0.0  & 2.5  & 0.0  & 3.5  & 0.0  & 1.1  & 0.0  & 2.8  & 0.0  & 1.1 \\

        OS-Atlas-4B      & 7.1  & 0.0  & 3.0  & 1.4  & 2.0  & 0.0  & 9.0  & 5.5  & 5.1  & 3.8  & 5.6  & 0.0  & 3.7 \\
        ShowUI-2B & 16.9 & 1.4  & 9.1  & 0.0  & 2.5  & 0.0  & 13.2 & 7.3  & 15.3 & 7.5  & 10.3 & 2.2  & 7.7 \\
        CogAgent-18B & $14.9$ & $0.7$  & $9.6$  & $0.0$  & $7.1$  & $3.1$  & $22.2$ & $1.8$  & $13.0$ & $0.0$  & $5.6$  & $0.0$  & 7.7 \\
        Aria-GUI & $16.2$ & $0.0$  & $23.7$ & $2.1$  & $7.6$  & $1.6$  & $27.1$ & $6.4$  & $20.3$ & $1.9$  & 4.7  & $0.0$  & 11.3 \\
        UGround-7B & 26.6 & $2.1$  & $27.3$ & $2.8$  & 14.2 & $1.6$  & $31.9$ & $2.7$  & $31.6$ & $11.3$ & 17.8 & $0.0$  & 16.5 \\
        OS-Atlas-7B & 33.1 & $1.4$  & 28.8 & $2.8$  & $12.2$ & 4.7  & 37.5 & $7.3$  & 33.9 & $5.7$  & 27.1 & 4.5  & 18.9 \\

        \midrule
        \multicolumn{12}{l}{\textbf{Generalist Models or Training-free Methods}  \,      \raisebox{-0.3ex}{\includegraphics[height=1em]{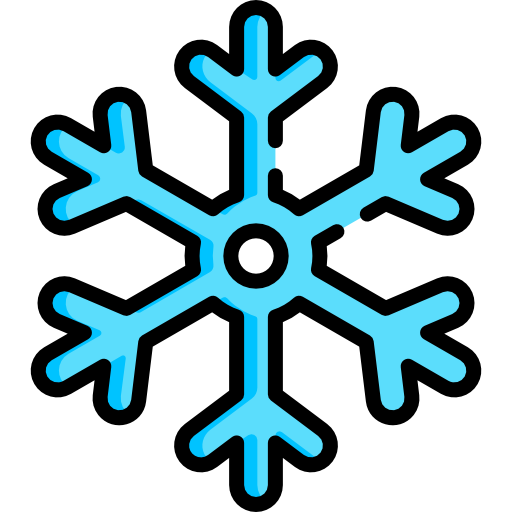}}} \\
        \midrule
        Qwen-VL-7B       & 0.0  & 0.0    & 0.0  & 0.0   & 0.0  & 0.0  & 0.7  & 0.0   & 0.0  & 0.0   & 0.0  & 0.0  & 0.1 \\
        GPT-4o  & 1.3  & 0.0   & 1.0  & 0.0   & 2.0  & 0.0   & 2.1  & 0.0   & 1.1  & 0.0   & 0.0  & 0.0  & 0.8 \\
        
        Qwen2-VL-7B & 2.6  & 0.0   & 1.5  & 0.0   & 0.5  & 0.0   & 6.3  & 0.0  & 3.4  & 1.9  & 0.9  & 0.0  & 1.6 \\
        MiniCPM-V         & 7.1  & 0.0  & 2.0  & 0.0  & 4.1  & \textbf{1.6}  & 8.3  & 0.0  & 2.8  & 3.8  & 3.7  & 1.1  & 3.0 \\
               
        \midrule
        Qwen2.5-VL-3B & 18.8 & 1.4 & 16.7 & 1.4 & 8.1 & \textbf{1.6} & 20.8 & 5.5 & 24.3 & 1.9 & 16.8 & 3.4 & 11.6 \\
        
        \rowcolor{blue!10} Qwen2.5-VL-3B + VPSG & 24.7 & 2.1 & \textbf{20.2} & 2.1 & \textbf{10.2} & \textbf{1.6} & 21.5 & 5.5 & 26.6 & 5.7 & 15.9 & 1.1 & 13.3 \\
        
        \midrule
        Qwen2.5-VL-7B & 37.7 & \textbf{2.8} & 19.7 & 2.1 & 7.6 & \textbf{1.6} & \textbf{31.3} & 5.5 & 41.8 & 11.3 & 29.9 & \textbf{10.1} & 18.5 \\
        
        \rowcolor{blue!10} Qwen2.5-VL-7B + VPSG & \textbf{40.9} & 2.1 & 19.8 & \textbf{2.8} & 8.1 & \textbf{1.6} & 30.6 & \textbf{5.6} & \textbf{43.5} & \textbf{13.2} & \textbf{29.9} & \textbf{10.1} & \textbf{19.1} \\
        
        \bottomrule
    \end{tabular}
   }
   
 \caption{Screen-based grounding results on \textsc{Screenspot-Pro}. Each column reports the evaluation score (higher is better) for a category (\emph{Development}, \emph{Creative}, \emph{CAD}, \emph{Scientific}, \emph{Office}, \emph{OS}) split by UI type (\emph{Text}/\emph{Icon}); \emph{Avg} is the unweighted mean across all columns. Rows marked ``\,+~VPSG'' apply our test-time guidance to the same base model, isolating the effect of the method. The best result among generalist models or training-free methods are highlighted in \textbf{bold} font. } 
    \label{tab:main}
\end{table*}

%% file: sec/table/ablationSeed.tex
\begin{table}[t]
\centering
\normalsize 

\begin{minipage}[t]{0.45\textwidth}
\centering
\textbf{Qwen2.5-VL-3B}

\begin{tabular}{l>{\hspace{12pt}}c>{\hspace{12pt}}c}
\toprule
\textbf{Setting} & \textbf{Avg} & $\Delta$ \\
\midrule
\rowcolor{black!7} VPSG & 13.3 & --\\
w/o Seeds aggregation & 13.0 & $\downarrow$ 0.3\\
w/o Coefficient decay & 11.9 & $\downarrow$ 1.4\\
\bottomrule
\end{tabular}
\end{minipage}
\vspace{5pt}
\begin{minipage}[t]{0.45\textwidth}
\centering
\textbf{Qwen2.5-VL-7B}

\begin{tabular}{l>{\hspace{12pt}}c>{\hspace{12pt}}c}
\toprule
\textbf{Setting} & \textbf{Avg} & $\Delta$ \\
\midrule
\rowcolor{black!7} VPSG & 19.1 & --\\
w/o Seeds aggregation & 18.6 & $\downarrow$ 0.5\\
w/o Coefficient decay & 18.2 & $\downarrow$ 0.9\\
\bottomrule
\end{tabular}
\end{minipage}
\caption{Ablation study results. The analyzed method components include:
(i) \emph{seeds aggregation}, the robust log-space aggregation of multiple PE-shuffled auxiliary routes (\textit{w/o seeds aggregation}: use a single seed, no aggregation); and
(ii) \emph{coefficient decay}, the position-aware geometric decay of the digit-only guidance weight with a reset at the first $y$ digit (\textit{w/o coefficient decay}: use a constant guidance weight across digits).}
\label{tab:ablation}
\vspace{-5pt}
\end{table}

%% file: sec/5conclusion.tex
\section{Conclusion}
We introduced Vision-PE Shuffle Guidance (VPSG) to rectify coordinate drift in multimodal large language models. Our causal analysis reveals that high-resolution inputs trigger directional and non-random biases. VPSG addresses this issue by using shuffled positional encodings as negative evidence to suppress spurious numeric priors. The method incorporates multi-seed aggregation and position-aware coefficient decay to stabilize the output. Evaluation on the ScreenSpot-Pro benchmark confirms consistent performance gains without any fine-tuning or architectural changes. These results highlight the critical role of robust positional encoding in fine-grained spatial reasoning. VPSG provides a practical solution for grounding and coordinate-sensitive tasks in future vision-language systems.

%% file: sec/appendix.tex
\section{Use of Large Language Models}
Large language models (LLMs) were used solely for language polishing and minor editorial assistance (e.g., grammar, wording, and clarity). 
They were not involved in the conception of research ideas, design of experiments, data analysis, or interpretation of results. 
All scientific content, methods, and conclusions were developed independently by the authors.

\section{Proof of Proposition (VPSG token guidance)}
\label{app:vpsg-proof}
\paragraph{Setup.}
Let $\mathcal{V}$ be the vocabulary and $\mathcal{D}\subset\mathcal{V}$ the digit subset. 
At decoding step $t$, denote by $p_A(v\mid c_t)$ the position-conditioned (main-route) distribution and by $p_B(v)$ the position-\emph{un}conditioned reference obtained from PE-shuffled auxiliary routes. 
Write $\ell_A(v\mid c_t)=\log p_A(v\mid c_t)$ and $\tilde{\ell}_B(v)=\log p_B(v)$. 
We assume:
(i) the same logits processors are applied to both routes before any guidance;
(ii) guidance acts only at the final-layer logits; and 
(iii) decoding is greedy, i.e., depends on the $\arg\max$ over logits.

\paragraph{CFG form.}
Consider the (digit-only) classifier-free guidance (CFG) mixing:
\begin{equation}\label{eq:cfg}
p_{\mathrm{CFG}}(v\mid c_t)\;\propto\;
\begin{cases}
\dfrac{p_A(v\mid c_t)^{\,1+\lambda_t}}{p_B(v)^{\,\lambda_t}}, & v\in\mathcal{D},\\[6pt]
p_A(v\mid c_t), & v\notin\mathcal{D},
\end{cases}
\end{equation}
with normalization constant $Z_{\mathrm{CFG}}(c_t)$ implicit over $v\in\mathcal{V}$.
Taking logs for $v\in\mathcal{D}$,
\begin{multline}\label{eq:cfglog}
\log p_{\mathrm{CFG}}(v\mid c_t) \\
= (1+\lambda_t)\,\ell_A(v\mid c_t)-\lambda_t\,\tilde{\ell}_B(v)-\log Z_{\mathrm{CFG}}(c_t).
\end{multline}

\paragraph{Digit-only positive affine rescaling.}
Define, for $v\in\mathcal{D}$,
\begin{multline}\label{eq:affine}
\psi_t(v) = \frac{1}{1+\lambda_t}\Big(\log p_{\mathrm{CFG}}(v\mid c_t)+\log Z_{\mathrm{CFG}}(c_t)\Big) \\
= \ell_A(v\mid c_t) - \underbrace{\frac{\lambda_t}{1+\lambda_t}}_{\alpha_t}\,\tilde{\ell}_B(v).
\end{multline}
For $v\notin\mathcal{D}$, keep $\psi_t(v)=\log p_A(v\mid c_t)$ (no change).

\begin{lemma}[Argmax invariance under positive affine transforms]
Let $S\subseteq\mathcal{V}$ and $a>0$, $b\in\mathbb{R}$. 
For any scores $\{u(v)\}_{v\in S}$, $\arg\max_{v\in S} u(v)=\arg\max_{v\in S} \{a\,u(v)+b\}$. 
\end{lemma}

\begin{proof}
For $a>0$, $u(v_1)\ge u(v_2)\iff a\,u(v_1)+b\ge a\,u(v_2)+b$.

Applying the lemma to \eqref{eq:cfglog} with $a=\tfrac{1}{1+\lambda_t}$ and $b=\tfrac{\log Z_{\mathrm{CFG}}(c_t)}{1+\lambda_t}$ \emph{restricted to $v\in\mathcal{D}$} shows that
\begin{equation}
\begin{aligned}
&\arg\max_{v\in\mathcal{D}} \log p_{\mathrm{CFG}}(v\mid c_t) = \arg\max_{v\in\mathcal{D}} \psi_t(v) \\
&\quad = \arg\max_{v\in\mathcal{D}} \big(\ell_A(v\mid c_t)-\alpha_t\,\tilde{\ell}_B(v)\big).
\end{aligned}
\end{equation}

Since $p_{\mathrm{CFG}}(v\mid c_t)=p_A(v\mid c_t)$ for $v\notin\mathcal{D}$ by \eqref{eq:cfg}, the combined $\arg\max$ under CFG equals that under the piecewise score 
\begin{equation}
s(v) = 
\begin{cases}
\ell_A(v\mid c_t)-\alpha_t\tilde{\ell}_B(v), & v\in\mathcal{D},\\[3pt]
\ell_A(v\mid c_t), & v\notin\mathcal{D},
\end{cases}
\end{equation}
which is the VPSG ``negative-evidence'' scoring form. 

Moreover, \eqref{eq:affine} yields the parameter mapping:
\begin{equation}
\boxed{\alpha_t = \frac{\lambda_t}{1+\lambda_t}} \quad \Longleftrightarrow \quad \boxed{\lambda_t = \frac{\alpha_t}{1-\alpha_t}}.
\end{equation}
\end{proof}
\paragraph{Equivalence under greedy decoding.}
Greedy decoding selects $\hat v_t=\arg\max_{v\in\mathcal{V}} \log p_{\mathrm{CFG}}(v\mid c_t)$.
By the lemma and the piecewise definition above, the same $\hat v_t$ is obtained by maximizing $s(v)$, because:
(i) on digits we used a positive affine transform of $\log p_{\mathrm{CFG}}$; 
(ii) on non-digits the two forms coincide; and 
(iii) both are compared in the same joint candidate set $\mathcal{V}$.
Therefore CFG \eqref{eq:cfg} and VPSG scoring $s(v)$ are \emph{decision-equivalent} under greedy decoding.

\paragraph{Normalization and distributional form.}
If a normalized distribution is desired, define
\begin{equation}
p_{\mathrm{VPSG}}(v\mid c_t) = \frac{\exp(s(v))}{\sum_{u\in\mathcal{V}}\exp(s(u))}.
\end{equation}
This coincides with \eqref{eq:cfg} up to the digit-only affine rescaling leading to the same $\arg\max$, and yields the proposition's statement:
\begin{equation}
\begin{aligned}
&p_{\mathrm{VPSG}}(v\mid c_t) \propto \\
&\begin{cases}
\exp\bigl(\ell_A(v\mid c_t)-\alpha_t\tilde{\ell}_B(v)\bigr), & v\in\mathcal{D},\\[3pt]
p_A(v\mid c_t), & v\notin\mathcal{D}.
\end{cases}
\end{aligned}
\end{equation}

\paragraph{Remarks on assumptions.}
(1) \emph{Same logits processors:} ensures that when $\alpha_t\to 0$ (or $\lambda_t\to 0$) the VPSG rule recovers the baseline exactly.
(2) \emph{Final-layer intervention:} guarantees that the affine transformation does not change any upstream normalization.
(3) \emph{Greedy decoding:} makes decision-equivalence depend only on $\arg\max$; for sampling or beam search, the same mapping holds at the score level, but selection statistics may also depend on temperature/length penalties (which can still be shared across routes).


\section{Details about dataset ScreenSpot-Pro}
\label{app:dataset}
We show categories and UI-type counts in Screenspot-Pro in~\Cref{tab:screenspot_counts}.
\begin{table}
\centering
\small
\begin{tabular}{lccc}
\toprule
\textbf{Group} & \textbf{Text} & \textbf{Icon} & \textbf{Total} \\
\midrule
CAD        & 197 & 64  & 261 \\
Creative   & 198 & 143 & 341 \\
Dev        & 154 & 145 & 299 \\
OS         & 107 & 89  & 196 \\
Office     & 177 & 53  & 230 \\
Scientific & 144 & 110 & 254 \\
\midrule
\textbf{All} & \textbf{977} & \textbf{604} & \textbf{1581} \\
\bottomrule
\end{tabular}
\caption{Category and UI-type counts in Screenspot-Pro.}
\label{tab:screenspot_counts}
\end{table}
\section{Algorithm of VPSG}
The complete algorithm is shown in \Cref{alg:vpsg}.
\label{app:algo}
\begin{algorithm}[!t]
\caption{Vision-PE Shuffle Guidance (VPSG)}
\label{alg:vpsg}
\begin{algorithmic}[1]
\Require Image $I$, prompt $q$, base model $\mathcal{M}$, seeds $\{s_1,\dots,s_S\}$, base coefficient $\alpha$, decay factor $0<\mathrm{decay}<1$
\Ensure Coordinate prediction $\hat y=[x,y]$

\State \textbf{Main route (position-conditioned):}
\State Run $\mathcal{M}$ on $(I,q)$ with normal positional encodings (PEs) to obtain token distribution $p_A(v\mid c_t)$.

\State \textbf{Auxiliary routes (position-unconditioned):}
\For{each seed $s$}
    \State Shuffle PEs and run $\mathcal{M}$ to get $p_B^{(s)}(v)$.
\EndFor
\State Aggregate in log-space:
\[
\tilde{\ell}_B(v) \gets \frac{1}{S}\sum_{s=1}^S \log p_B^{(s)}(v).
\]

\State \textbf{FSM state tracking:}
\State Use a finite-state machine aligned to the $[x,y]$ template to determine whether the model is decoding a digit in $x$ or $y$.

\State \textbf{Coefficient scheduling:}
\If{decoding the $k_x$-th digit of $x$}
    \State $\alpha_t \gets \alpha\cdot\mathrm{decay}^{\,k_x-1}$
\ElsIf{decoding the first digit of $y$}
    \State $\alpha_t \gets \alpha$
\ElsIf{decoding the $k_y$-th digit of $y$}
    \State $\alpha_t \gets \alpha\cdot\mathrm{decay}^{\,k_y-1}$
\EndIf

\State \textbf{Negative-evidence scoring:}
\For{each token $v$}
    \State
    \[
    s(v) \gets
    \begin{cases}
    \log p_A(v\mid c_t) - \alpha_t \tilde{\ell}_B(v), & v \in \mathcal{D},\\[4pt]
    \log p_A(v\mid c_t), & v \notin \mathcal{D}.
    \end{cases}
    \]
\EndFor

\State \textbf{Token selection:}
\State Choose $\hat v_t \gets \arg\max_v s(v)$, append to output, and advance FSM.

\State \textbf{Termination:}
\State Repeat Steps 3--6 until EOS. Decode tokens into coordinates $\hat y$.
\end{algorithmic}
\end{algorithm}

\section{Expected distance in the unit square}
\label{app:proof1}

\begin{claim}
If $X=(X_1,X_2)$ and $Y=(Y_1,Y_2)$ are independent and uniformly distributed on $[0,1]^2$, then
\begin{equation}
\mathbb{E}\big[\|X-Y\|_2\big]
=\frac{2+\sqrt{2}+5\ln(1+\sqrt{2})}{15}
\approx 0.5214.
\end{equation}
\end{claim}

\begin{proof}
Let $U=|X_1-Y_1|$ and $V=|X_2-Y_2|$. For $u,v\in[0,1]$, the joint density of $(U,V)$ is $f_{U,V}(u,v)=4(1-u)(1-v)$, as $U,V$ are i.i.d. triangular variables with $f_U(u)=2(1-u)$. The Euclidean distance is $R=\sqrt{U^2+V^2}$. Hence:
\begin{equation}
\mathbb{E}[R] = \int_0^1\!\!\int_0^1 4\sqrt{u^2+v^2}(1-u)(1-v)\,du\,dv.
\end{equation}
Switching to polar coordinates ($u=r\cos\theta, v=r\sin\theta$), the boundary $0 \le r \le r_{\max}(\theta)$ where $r_{\max}(\theta)=\min\{\sec\theta, \csc\theta\}$ gives:
\begin{equation}
\begin{aligned}
\mathbb{E}[R] = &4 \int_0^{\pi/2}\int_0^{r_{\max}(\theta)} r^2(1-r\cos\theta) \\
&\times (1-r\sin\theta)\,dr\,d\theta.
\end{aligned}
\end{equation}
Splitting the integral at $\theta=\pi/4$ where the boundary constraint changes:
\begin{equation}
\begin{aligned}
\mathbb{E}[R] &= 4\int_0^{\pi/4}\int_0^{\sec\theta} g(r,\theta)\,dr\,d\theta \\
&+ 4\int_{\pi/4}^{\pi/2}\int_0^{\csc\theta} g(r,\theta)\,dr\,d\theta,
\end{aligned}
\end{equation}
where $g(r,\theta) = r^2-r^3(\cos\theta+\sin\theta)+r^4\sin\theta\cos\theta$. Integrating in $r$ for a general upper limit $a$:
\begin{equation}
\int_0^{a} g(r,\theta) dr = \frac{a^3}{3}-\frac{\cos\theta+\sin\theta}{4}a^4+\frac{\sin\theta\cos\theta}{5}a^5.
\end{equation}
Substituting $a=\sec\theta$ for the first range $[0,\pi/4]$ yields:
\begin{equation}
\begin{aligned}
J_1(\theta) &= \frac{\sec^3\theta}{3} \\
&\quad - \frac{\cos\theta+\sin\theta}{4}\sec^4\theta \\
&\quad + \frac{\sin\theta\cos\theta}{5}\sec^5\theta \\
&= \sec^3\theta \left( \frac{1}{12} - \frac{1}{20}\tan\theta \right).
\end{aligned}
\end{equation}
Symmetrically, for $a=\csc\theta$ on $[\pi/4, \pi/2]$, we have $J_2(\theta) = \csc^3\theta(\frac{1}{12}-\frac{1}{20}\cot\theta)$. Thus:
\begin{equation}
\begin{aligned}
\mathbb{E}[R] &= 4\int_{0}^{\pi/4} \left(\frac{1}{12}\sec^3\theta-\frac{1}{20}\sec^3\theta\tan\theta\right)d\theta \\
&+ 4\int_{\pi/4}^{\pi/2} \left(\frac{1}{12}\csc^3\theta-\frac{1}{20}\csc^3\theta\cot\theta\right)d\theta.
\end{aligned}
\end{equation}
Using the antiderivatives $\int \sec^3\theta \, d\theta = \frac{1}{2}(\sec\theta\tan\theta + \ln(\sec\theta+\tan\theta))$ and $\int \sec^3\theta\tan\theta \, d\theta = \frac{1}{3}\sec^3\theta$, and noting the symmetry between the two $\theta$-ranges, the integral evaluates to:
\begin{equation}
\begin{aligned}
\mathbb{E}[R] &= \frac{1}{3}(\sqrt{2}+\ln(1+\sqrt{2})) - \frac{2}{15}(2\sqrt{2}-1) \\
&= \frac{2+\sqrt{2}+5\ln(1+\sqrt{2})}{15}.
\end{aligned}
\end{equation}
This completes the proof.
\end{proof}

\paragraph{Remark (Application to images)}
The constant $\mu_\square$ is the dispersion benchmark for a unit square. For arbitrary image sizes $(W,H)$, one can: (i) anisotropically rescale coordinates to $[0,1]^2$ before computing distances, or (ii) form a per-image Monte-Carlo null by sampling i.i.d. uniform points from $[0,W]\times[0,H]$ to estimate the baseline for that specific aspect ratio.

\section{More Experiment Result}
\subsection{distance analysis}
More details of distance analysis results are shown in~\Cref{tab:distance_stats_combined}
\begin{table}[]

\vspace{5pt}
\centering
\tiny
\begin{tabular}{lcccc}
\toprule
\multirow{2}{*}{Image size} &
\multicolumn{2}{c}{Qwen2.5-VL-3B} &
\multicolumn{2}{c}{Qwen2.5-VL-7B} \\
\cmidrule(lr){2-3}\cmidrule(lr){4-5}
& Normal & Shuffled & Normal & Shuffled \\
\midrule
1920$\times$1080 & 0.123 & 0.202 & 0.211 & 0.128 \\
2160$\times$1440 & 0.391 & 0.146 & 0.437 & 0.159 \\
2560$\times$1440 & 0.392 & 0.171 & 0.436 & 0.180 \\
2560$\times$1600 & 0.357 & 0.072 & 0.348 & 0.048 \\
2560$\times$1664 & 0.428 & 0.139 & 0.471 & 0.211 \\
2880$\times$1800 & 0.379 & 0.119 & 0.462 & 0.111 \\
2992$\times$1870 & 0.463 & 0.175 & 0.347 & 0.061 \\
3456$\times$2160 & 0.471 & 0.122 & 0.348 & 0.081 \\
3456$\times$2234 & 0.457 & 0.169 & 0.466 & 0.106 \\
3840$\times$1080 & 0.298 & 0.227 & 0.287 & 0.140 \\
3840$\times$2160 & 0.468 & 0.117 & 0.504 & 0.146 \\
5120$\times$1440 & 0.435 & 0.133 & 0.404 & 0.058 \\
5120$\times$2880 & 0.399 & 0.168 & 0.408 & 0.052 \\
6016$\times$3384 & 0.341 & 0.117 & 0.335 & 0.057 \\
\midrule
\textbf{Overall mean} & \textbf{0.402} & \textbf{0.164} & \textbf{0.435} & \textbf{0.162} \\
\bottomrule
\end{tabular}
\caption{
Comparison of mean pairwise Euclidean distances (diagonal-normalized) across image resolutions on \textsc{Screenspot-Pro}.  
Normal PEs exhibit significantly larger within-input distances, while shuffled PEs lead to much tighter clustering and systematic collapse of coordinate predictions, demonstrating the impact of positional information on spatial diversity.
}
\label{tab:distance_stats_combined}
\end{table}

\subsection{Case study with per-step logits.}
\begin{figure*}[t]
    \centering
    \includegraphics[width=\linewidth]{fig/screenshot_2024-10-27_21-07-29.png}
    \caption{
 An image case from dataset ScreenSpot-Pro (ppt\_windows/screenshot\_2024-10-27\_21-07-29.png)
    }
    \label{fig:casestudy}
\end{figure*}
To better illustrate how VPSG corrects coordinate prediction bias, we analyze a single case from \textsc{Screenspot-Pro} (\path{ppt_windows/screenshot_2024-10-27_21-07-29.png})~\Cref{fig:casestudy}.
The ground-truth bounding box center is $[659,857]$.  
The base model without guidance predicts $[1024,856]$, while VPSG successfully outputs the correct $[659,857]$.  
Table~\ref{tab:vpsg_logits_case} lists the top-10 logits probabilities at each decoding step.  
At the earliest $x$-digit steps, the uncorrected model shows a strong bias toward larger numbers (e.g., tokens ``1'' and ``0'' dominate), reflecting spurious numeric priors induced by missing or unreliable positional encodings.  
VPSG integrates negative evidence from multiple shuffled PE runs and systematically downweights these spurious peaks, allowing the true digit sequence to emerge and stabilizing the final $[x,y]$ prediction.

\begin{table}[t]

\vspace{5pt}
\centering
\scriptsize
\begin{tabular}{c|c|c|c}
\toprule
\textbf{Step} & \textbf{Token} & \textbf{VPSG Prob.} & \textbf{Base Prob.} \\
\midrule
1 & [  & 0.805 & 0.805 \\
2 & 6  & 0.243 & 0.210 \\
3 & 5  & 0.172 & 0.264 (0 highest) \\
4 & 9  & 0.221 & 0.140 (2 highest) \\
5 & ,  & 0.999 & 0.987 \\
6 & space & 0.999 & 0.976 \\
7 & 8  & 0.740 & 0.355 \\
8 & 5  & 0.589 & 0.642 (but 6/3 confused) \\
9 & 7  & 0.272 & 0.421 (6 highest) \\
10 & ] & 0.999 & 0.999 \\
11 & \textless eos\textgreater & 0.993 & 0.996 \\
\bottomrule
\end{tabular}
\caption{
Per-step top-1 logits probabilities for each decoded token in a representative example.  
The base model without guidance drifts toward spurious large-$x$ digits (e.g., ``1'', ``0'' early), yielding an incorrect coordinate $[1024,856]$.  
VPSG, by integrating negative evidence from multiple shuffled PE runs, suppresses these biased peaks and converges to the correct coordinate $[659,857]$.  
This example highlights how VPSG stabilizes numeric decoding and restores faithful spatial grounding.
}
\label{tab:vpsg_logits_case}
\end{table}

\subsection{Ablation: removing seeds aggregation.}
To evaluate the contribution of seeds aggregation in VPSG, we remove the multi-seed log-space aggregation and instead rely on a single randomly shuffled positional encoding as the auxiliary route. 
Table~\ref{tab:ablation_seed} reports the detailed category- and type-level accuracies (percentage of correct predictions) for both Qwen2.5-VL-3B and Qwen2.5-VL-7B models.  
Without aggregation, accuracy drops across almost all groups and UI-types, confirming that a single random shuffle provides only a noisy sample of the underlying position-unconditioned bias distribution and cannot capture its full expectation.  
This validates the theoretical claim that multi-seed aggregation approximates the expected bias prior and yields more stable and accurate guidance.

\begin{table}[t]

\vspace{5pt}
\centering
\scriptsize
\begin{tabular}{lcccc}
\toprule
\multirow{2}{*}{Category / UI type} & \multicolumn{2}{c}{\textbf{3B}} & \multicolumn{2}{c}{\textbf{7B}} \\
\cmidrule(lr){2-3}\cmidrule(lr){4-5}
 & Text (\%) & Icon (\%) & Text (\%) & Icon (\%) \\
\midrule
CAD         & 10.66 & 1.56 & 6.60  & 1.56 \\
Creative    & 18.69 & 2.10 & 19.70 & 2.80 \\
Dev         & 22.73 & 2.07 & 41.56 & 3.45 \\
OS          & 15.89 & 2.25 & 28.97 &12.36 \\
Office      & 29.38 & 5.66 & 41.81 &13.21 \\
Scientific  & 20.14 & 2.73 & 27.78 & 4.55 \\
\midrule
Overall     & 19.55 & 2.48 & 26.71 & 5.46 \\
\bottomrule
\end{tabular}
\caption{
Accuracy (\%) of VPSG without seeds aggregation across categories and UI types.  
Both model sizes show clear drops compared with the full VPSG (see main text~\Cref{tab:ablation}): for instance, the 3B model drops from $13.3\%$ overall to $13.0\%$, and the 7B model from $19.1\%$ to $18.6\%$.  
Losses are consistent across text and icon settings, supporting the view that multi-seed aggregation provides a faithful estimate of the expected position-unconditioned bias and stabilizes the guidance effect.
}
\label{tab:ablation_seed}
\end{table}
\begin{table*}[!t]
\centering

\vspace{5pt}
\label{tab:predcoord-freq}
\resizebox{\textwidth}{!}{
\begin{tabular}{lrr lrr lrr lrr}
\toprule
\multicolumn{3}{c}{\textbf{Qwen2.5-VL-7B (normal PE)}} &
\multicolumn{3}{c}{\textbf{Qwen2.5-VL-3B (normal PE)}} &
\multicolumn{3}{c}{\textbf{Qwen2.5-VL-3B (shuffled PE)}} &
\multicolumn{3}{c}{\textbf{Qwen2.5-VL-7B (shuffled PE)}}\\
\cmidrule(lr){1-3}\cmidrule(lr){4-6}\cmidrule(lr){7-9}\cmidrule(lr){10-12}
\textbf{Rank} & \textbf{Number} & \textbf{Freq} &
\textbf{Rank} & \textbf{Number} & \textbf{Freq} &
\textbf{Rank} & \textbf{Number} & \textbf{Freq} &
\textbf{Rank} & \textbf{Number} & \textbf{Freq}\\
\midrule
1  & 1024 & 296 & 1  & 1024 & 397 & 1  & 1024 & 591 & 1  & 1024 & 902 \\
2  & 105  & 54  & 2  & 1056 & 82  & 2  & 1056 & 426 & 2  & 1234 & 582 \\
3  & 10   & 35  & 3  & 356  & 73  & 3  & 568  & 184 & 3  & 567  & 580 \\
4  & 2048 & 26  & 4  & 35   & 44  & 4  & 1000 & 182 & 4  & 672  & 270 \\
5  & 2058 & 26  & 5  & 36   & 39  & 5  & 238  & 159 & 5  & 368  & 266 \\
6  & 2016 & 25  & 6  & 10   & 36  & 6  & 512  & 141 & 6  & 384  & 162 \\
7  & 1940 & 23  & 7  & 1234 & 28  & 7  & 560  & 141 & 7  & 200  & 41  \\
8  & 100  & 23  & 8  & 105  & 27  & 8  & 1052 & 128 & 8  & 36   & 33  \\
9  & 200  & 21  & 9  & 102  & 25  & 9  & 200  & 121 & 9  & 38   & 27  \\
10 & 1056 & 20  & 10 & 1048 & 23  & 10 & 248  & 119 & 10 & 896  & 26  \\
\bottomrule
\end{tabular}}
\vspace{4pt}
\footnotesize
\caption{Top-10 most frequent numbers appearing in prediction results for Qwen2.5-VL-3B and Qwen2.5-VL-7B under normal and shuffled positional encodings.}
\label{tab:top10}
\end{table*}
\subsection{More details about bias analysis}
\Cref{tab:top10} reports the ten most frequent individual numbers across all $[x, y]$ coordinate predictions. 
Normal PE = standard positional encodings; shuffled PE = visual positional encodings randomly shuffled at inference time. 
All counts reflect total occurrences of a number as either the $x$ or $y$ component of predicted coordinates.